\PassOptionsToPackage{table}{xcolor}
\documentclass[sigconf,screen]{acmart}
\usepackage{xurl}
\usepackage{dblfloatfix} 

\usepackage{amsmath}            
\usepackage{booktabs}           
\usepackage{multirow}           
\usepackage{tabularx}           
\usepackage{pifont}             
\usepackage{balance}            

\newcommand{\cmark}{\ding{51}}
\newcommand{\xmark}{\ding{55}}

\title{BadmintonGRF: A Multimodal Dataset and Benchmark for Markerless Ground Reaction Force Estimation in Badminton}

\author{Kuoye Niu}
\affiliation{
  \institution{Beijing Sport University}
  \city{Beijing}
  \country{China}
}
\affiliation{
  \institution{Zhejiang University}
  \city{Hangzhou}
  \country{China}
}
\email{0924632@zju.edu.cn}

\author{Jianwei Li}
\affiliation{
  \institution{Beijing Sport University}
  \city{Beijing}
  \country{China}
}
\email{jianwei@bsu.edu.cn}

\author{Shengze Cai}
\affiliation{
  \institution{Zhejiang University}
  \city{Hangzhou}
  \country{China}
}
\email{shengze_cai@zju.edu.cn}

\author{Yong Ma}
\affiliation{
  \institution{Wuhan Sports University}
  \city{Wuhan}
  \country{China}
}
\email{mayong@whsu.edu.cn}

\author{Mengyao Jia}
\affiliation{
  \institution{Wuhan Sports University}
  \city{Wuhan}
  \country{China}
}
\email{2023420024@whsu.edu.cn}

\author{Lishun Shen}
\affiliation{
  \institution{Zhejiang Gongshang University}
  \city{Hangzhou}
  \country{China}
}
\email{shenlishun@mail.zjgsu.edu.cn}

\author{Zhenheng Zhang}
\affiliation{
  \institution{Zhejiang University}
  \city{Hangzhou}
  \country{China}
}
\email{zhzh@zju.edu.cn}

\author{Yuxin Peng}
\affiliation{
  \institution{Zhejiang University}
  \city{Hangzhou}
  \country{China}
}
\email{pyxpeng@zju.edu.cn}

\author{Xian Song}
\authornote{Corresponding author.}
\affiliation{
  \institution{Zhejiang University}
  \city{Hangzhou}
  \country{China}
}
\email{sx1993@zju.edu.cn}

\ccsdesc[500]{Computing methodologies~Computer vision}
\ccsdesc[500]{Applied computing~Physical sciences and engineering}
\ccsdesc[300]{Information systems~Multimedia databases}

\keywords{dataset, multimodal learning, computer vision, ground reaction force, badminton, biomechanics, sports video}

\acmConference[MM '26]{ACM Multimedia 2026}{November 10--14, 2026}{Rio de Janeiro, Brazil}
\acmBooktitle{Proceedings of the 34th ACM International Conference on Multimedia (MM '26)}
\acmDOI{10.1145/nnnnnnn.nnnnnnn}

\newcommand{\rateHz}[1]{\mbox{#1~Hz}}
\newcommand{\BadmintonGRFGitHubBase}{https://github.com/KenyaNiu/BadmintonGRF}
\newcommand{\BadmintonGRFRepo}{\url{\BadmintonGRFGitHubBase}}
\newcommand{\BadmintonGRFRepoShort}{\href{\BadmintonGRFGitHubBase}{project page}}
\newcommand{\BadmintonGRFSupplementary}{\emph{Supplementary Material}}
\newcommand{\QAabbrev}{QA}

\begin{document}
\raggedbottom

\begin{abstract}
Multimodal resources for non-periodic court sports with laboratory-grade sensing remain scarce: few publicly pair instrumented ground reaction force (GRF) with high-frame-rate multi-view video, limiting markerless load estimation in realistic training settings.
BadmintonGRF records eight synchronized RGB views at \(\sim\)120\,FPS, four Kistler force plates, and Vicon motion capture (C3D) without hardware genlock across modalities; alignment combines human-verified events, automated quality assurance, and per-camera time offsets with uncertainty metadata.
Tier~1 distributes pose, time-aligned GRF, metadata, and splits under CC BY-NC~4.0, enabling the primary benchmark without raw RGB or C3D; we report a Tier~1 task that maps 2D pose to GRF.
Tier~2 provides raw RGB and C3D under controlled access for studies that require appearance or full kinematics.
The public release contains 17{,}425 impact-segment archives in the 10-subject benchmark tree (\(156\) instrumented trials; raw multi-view RGB alone exceeds \(1\,\mathrm{TB}\)); benchmark loader gates retain 12{,}867 view-specific instances and 1{,}732 unique impacts after multi-view deduplication.
We are not aware of prior \emph{public} badminton corpora that combine this sensing layout with audited video--GRF alignment for impact-centric GRF estimation.
We distribute preprocessing code, leave-one-subject-out splits, ten reference baselines, and optional late fusion (one deterministic test-time pass per instance; no test-time augmentation), with a within-trial diagnostic in \BadmintonGRFSupplementary{}.
\end{abstract}

\maketitle

\section{Introduction}
\begin{figure}[!t]
  \centering
  \includegraphics[width=0.98\linewidth]{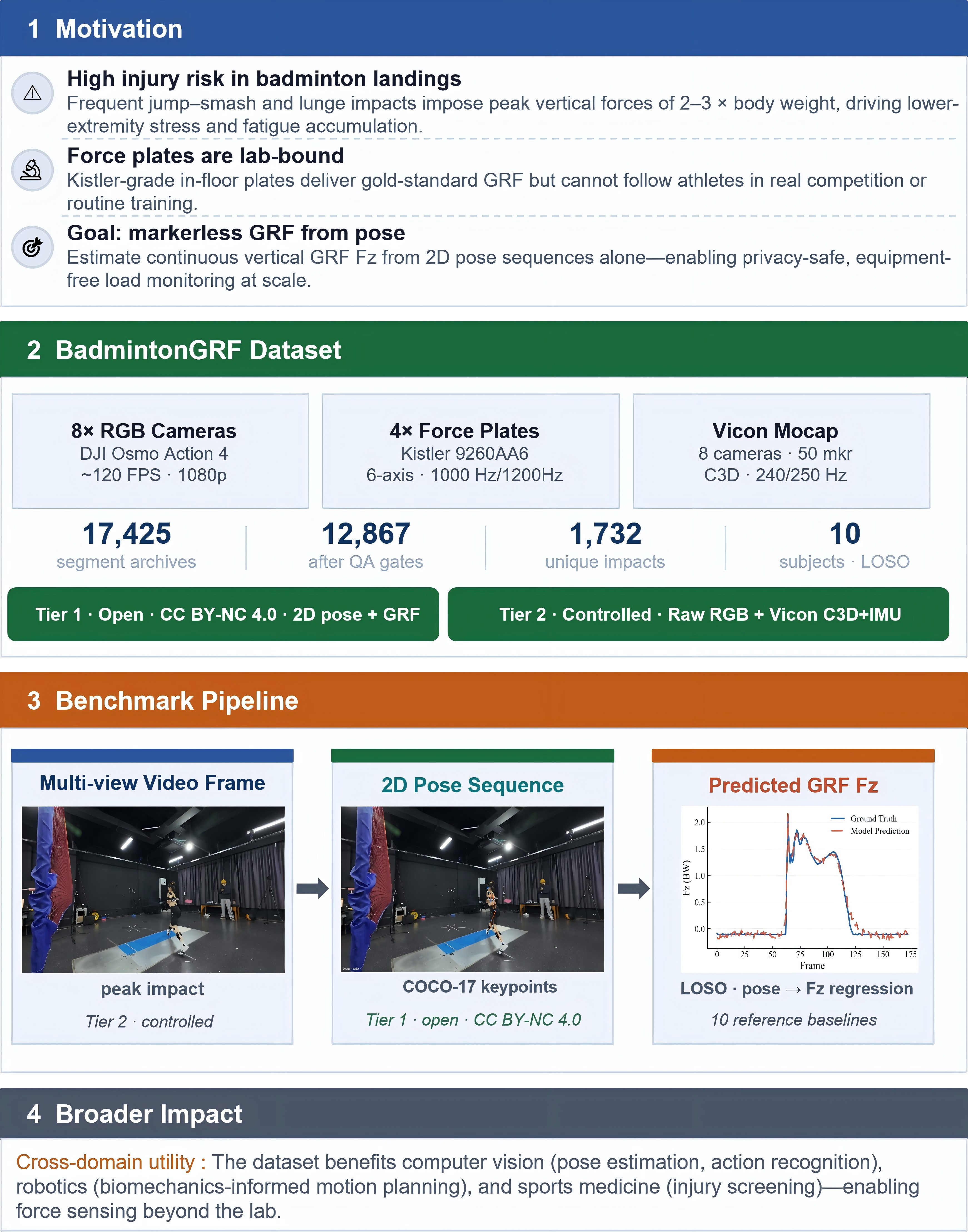}%
  \caption{BadmintonGRF at a glance: impact-centric badminton movement, synchronized multi-view video and pose with laboratory sensing, and benchmark targets for GRF regression. Each impact window predicts body-weight (BW) normalized vertical force \((F_z)\) from visual inputs under fixed alignment and protocol metadata.}
  \label{fig:overview}
  \Description{Teaser figure of BadmintonGRF: an impact-centric learning setup with synchronized multi-view video and 2D pose streams aligned to laboratory vertical ground reaction force targets, plus benchmark outputs for GRF regression.}
\end{figure}

Ground reaction forces (GRF) summarize external foot--ground loading and underpin workload monitoring, fatigue analysis, and lower-extremity injury risk assessment \cite{lamBiomechanicsLowerLimb2020,pardiwalaBadmintonInjuriesElite2020,huCorrelationLowerLimb2022,valldecabresEffectMatchFatigue2022}.
Markerless estimation typically chains pose estimation from video to predictive models of foot--ground forces; when calibrated on instrumented floors, it could extend load monitoring beyond sessions that require force plates.
Laboratory-grade GRF nonetheless depends on plates and tight cross-sensor timing, which is costly to deploy for continuous, unconstrained capture \cite{evansSynchronisedVideoMotion2024,uhlrichOpenCapHumanMovement2023,werlingAddBiomechanicsDatasetCapturing2025}.
BadmintonGRF contributes \emph{study-grade} multimodal capture to narrow that gap: high-rate multi-view video, Vicon C3D kinematics, and reference six-axis GRF with software video--GRF alignment (consumer RGB is not genlocked to the lab clock), human-verified event pairing, and systematic quality assurance (QA); an open leave-one-subject-out (LOSO) benchmark pairs privacy-preserving pose with time-aligned vertical \(F_z\) (Fig.~\ref{fig:overview}).

\paragraph{Why badminton and why now.}
Learning-based GRF estimation from video and pose has advanced \cite{mundt2022Estimating,ishida2024Estimation,hossain2025Knowledge}, yet evaluation still centers on periodic gait or scripted jumps.
Competitive badminton is \emph{intermittent}: rapid direction changes and frequent landings produce irregular foot--ground loading compared with steady locomotion.
Those demands link directly to match workload and lower-limb stress \cite{lamBiomechanicsLowerLimb2020,pardiwalaBadmintonInjuriesElite2020,huCorrelationLowerLimb2022,tongEffectsAnkleDorsiflexor2023,valldecabresEffectMatchFatigue2022}.
GRF estimation under such variability is difficult but closer to on-court loading than homogeneous stride datasets.
Public resources that combine instrumented GRF, high-rate multi-view video, and a reproducible badminton impact benchmark are nevertheless rare.
BadmintonGRF targets that void with plate-aligned recordings, deterministic loaders, and a fixed evaluation recipe.

\paragraph{Positioning vs. existing badminton datasets.}
FineBadminton emphasizes fine-grained semantic understanding of rallies through hierarchical labels and an MLLM-assisted annotation workflow \cite{heFineBadmintonMultiLevelDataset2025}.
BadmintonGRF complements such corpora with \emph{biomechanical ground truth}: time-synchronized force-plate GRF and mocap suited to \emph{markerless} load estimation from video or pose, together with protocol-stratified metadata and a fixed benchmark recipe.
The cohort comprises \emph{national second-tier or higher} athletes in active training; besides RGB and GRF, we distribute C3D and optional inertial measurement unit (IMU) streams with documented limitations, organized around standard badminton technique progressions.
Framed for the Dataset Track, our emphasis is on data quality, cross-modal alignment, and evaluation that others can reproduce verbatim.

\paragraph{Contributions.}
Following Dataset Track expectations, BadmintonGRF foregrounds the resource, documentation, and evaluation protocol; reference models supply reproducibility anchors rather than performance claims relative to prior art.
Our main contributions are:
\begin{enumerate}
  \item \textbf{Study-grade multimodal capture}: eight-view \(\sim\)120\,FPS RGB, four Kistler force plates, Vicon C3D, and fatigue-aware protocol tags for non-periodic badminton impacts---among the first public releases of this combination, to our knowledge.
  \item \textbf{Software alignment without genlock}: a human-in-the-loop pipeline with automated QA that reaches roughly frame-level alignment between consumer RGB and laboratory sensing in typical lab configurations.
  \item \textbf{Packaged benchmark}: 17{,}425 exported impact segments (12{,}867 after loader gates; 1{,}732 unique impacts) with deterministic splits and loaders.
  \item \textbf{Transparent evaluation}: fixed LOSO folds, four complementary metrics, ten documented baselines, and scripts that reduce unreported implementation variance across sites.
\end{enumerate}
Overall we stress \emph{resource quality and reuse}: instrumented ground truth, explicit alignment and \QAabbrev{} practice, and benchmark code paths that can be shared without hidden preprocessing choices.

\paragraph{Relevance to ACM Multimedia.}
Beyond sports biomechanics, the corpus supports multimedia research on heterogeneous clocks without genlock, fixed multi-view geometry, occluded pose-to-dynamics modeling, domain shift between lab and training venues, and subject transfer---topics aligned with vision and embodied media.
A tiered design separates privacy-preserving Tier~1 pose and GRF from Tier~2 raw RGB and C3D under controlled access (Sec.~\ref{sec:data_access}).
The headline benchmark (Table~\ref{tab:benchmark_all}) uses Tier~1 alone so results are comparable across laboratories without handling restricted video.
The paper first documents \emph{resource construction} (Sec.~\ref{sec:related}--Sec.~\ref{sec:dataset_stats}), then \emph{benchmark use} (Sec.~\ref{sec:experiments}).

\section{Related Work}
\label{sec:related}
We summarize badminton corpora, laboratory-synchronized biomechanics data, and pose-to-GRF learning, then state what BadmintonGRF adds for intermittent court sports.
\subsection{Badminton Datasets and Understanding Benchmarks}
\begin{table}[t]
  \centering
  \caption{Public badminton and related biomechanics resources versus BadmintonGRF. \textbf{MV}: multi-view video; \textbf{MC}: optical motion capture; \textbf{GRF}: laboratory in-floor force plates (\xmark{} indicates wearable pressure only, not instrumented flooring); \textbf{Ftg}: fatigue or protocol stage metadata.}
  \label{tab:modality_compare}
  \footnotesize
  \begin{minipage}{\columnwidth}\sloppy
  \setlength{\tabcolsep}{2.1pt}
  \renewcommand{\arraystretch}{1.1}
  \begin{tabularx}{\columnwidth}{@{}>{\raggedright\arraybackslash}p{1.55cm} >{\raggedright\arraybackslash}X >{\raggedright\arraybackslash}p{1.45cm} cccc@{}}
    \toprule
    \textbf{Dataset} & \textbf{Focus} & \textbf{Scale} & \multicolumn{4}{c@{}}{\textbf{Sensing}} \\
    \cmidrule(l{2pt}r{2pt}){4-7}
     & & & \textbf{M} & \textbf{C} & \textbf{G} & \textbf{F} \\
    \midrule
    ShuttleNet~\cite{wangShuttleNetPositionAwareFusion2022} & Stroke forecasting; rally anticipation & 75 matches & \xmark & \xmark & \xmark & \xmark \\
    ShuttleSet~\cite{wangShuttleSetHumanAnnotatedStrokeLevel2023} & Singles stroke tactics & 44 matches & \xmark & \xmark & \xmark & \xmark \\
    \shortstack[l]{Fine\\Badminton}~\cite{heFineBadmintonMultiLevelDataset2025} & Hier.\ semantics; MLLM tasks & 120 matches & \xmark & \xmark & \xmark & \xmark \\
    \shortstack[l]{MultiSense\\Badminton}~\cite{seongMultiSenseBadmintonWearableSensor2024} & Wearable pressure \& perf. & 7{,}763 swings & \xmark & \xmark & \xmark & \xmark \\
    BioCV~\cite{evansSynchronisedVideoMotion2024} & Walk/run/CMJ/hop (lab) & 600 instr.\ trials & \cmark & \cmark & \cmark & \xmark \\
    \cmidrule[\heavyrulewidth](l{2pt}r{2pt}){1-7}
    \shortstack[l]{\textbf{Badminton}\\\textbf{GRF}} & \textbf{Impacts; GRF bench (LOSO)} & \shortstack[l]{\textbf{17,425} seg.;\\\textbf{1,732} uniq.$^{\dagger}$} & \cellcolor{black!4}\cmark & \cellcolor{black!4}\cmark & \cellcolor{black!4}\cmark & \cellcolor{black!4}\cmark \\
    \bottomrule
  \end{tabularx}
  \vspace{2pt}
  {\setlength{\baselineskip}{10pt}%
  \noindent\textbf{M}/\textbf{C}/\textbf{G}/\textbf{F} abbreviate \textbf{MV}, \textbf{MC}, \textbf{GRF}, and \textbf{Ftg} from the caption.
  $^{\dagger}$Counts are on-disk impact windows and unique $(\text{trial},\text{impact})$ pairs after quality gates (Sec.~\ref{sec:dataset_stats}).
  Tier~1 publicly lists ten benchmark subjects drawn from seventeen instrumented athletes.
  Additional cross-dataset rows: \BadmintonGRFSupplementary{}.}
  \end{minipage}
\end{table}
Prior badminton releases foreground tactics and semantics rather than instrumented GRF, mocap, and \(\sim\)120\,FPS multi-view coverage for landing impacts \cite{wangShuttleSetHumanAnnotatedStrokeLevel2023,heFineBadmintonMultiLevelDataset2025,seongMultiSenseBadmintonWearableSensor2024}.
Table~\ref{tab:modality_compare} compares scale, task focus, and sensing modalities (\textbf{M}/\textbf{C}/\textbf{G}/\textbf{F} = MV/MC/GRF/protocol tags).
\textbf{BioCV} \cite{evansSynchronisedVideoMotion2024} offers a controlled multi-sensor stack but targets generic locomotion rather than badminton-specific footwork.
Additional cross-sport rows appear in \BadmintonGRFSupplementary{}.

\subsection{Multi-view Synchronized Biomechanics and GRF Estimation}
Multi-view sensing eases occlusion and viewpoint turnover on court \cite{bragagnolo2025Multiview,cao2025KeypointNet}.
\textbf{BioCV} \cite{evansSynchronisedVideoMotion2024} is the nearest counterpart in instrumentation but emphasizes stereotyped gait and jumps instead of badminton footwork or fatigue stratification.
\textbf{AddBiomechanics} \cite{werlingAddBiomechanicsDatasetCapturing2025} scales physics-informed modeling yet does not supply our badminton protocol, per-trial alignment review, or impact-window task definition.
BadmintonGRF responds with sport-specific capture plus end-to-end alignment, segmentation, and release tooling.

\subsection{From Pose/Video to GRF: Tracking and Temporal Modeling}
Non-contact GRF estimation from 2D pose or video is an active line of work \cite{ishida2024Estimation,mundt2022Estimating,hossain2025Knowledge,johnsonOnfieldPlayerWorkload2019,kim2025Estimation,gaoAutomatedRecognitionAsymmetric2023,bogaertPredictingVerticalGround2024}.
Strong pose and tracking backbones are available off the shelf \cite{zhangByteTrackMultiobjectTracking2022,fangAlphaPoseWholeBodyRegional2023,linFeaturePyramidNetworks2017,linMicrosoftCOCOCommon2014,caoRealtimeMultiPerson2017,bazarevskyBlazePose2020,songHumanPoseEstimation2021,zhou2023Efficient}.
A community missing piece is a badminton benchmark with plate-synchronized targets, unified quality control, and a fixed LOSO protocol.
BadmintonGRF supplies that packaging alongside single-view and late-fusion reference models.

\section{BadmintonGRF Dataset}
\begin{figure}[!t]
  \centering
  \includegraphics[width=0.86\linewidth]{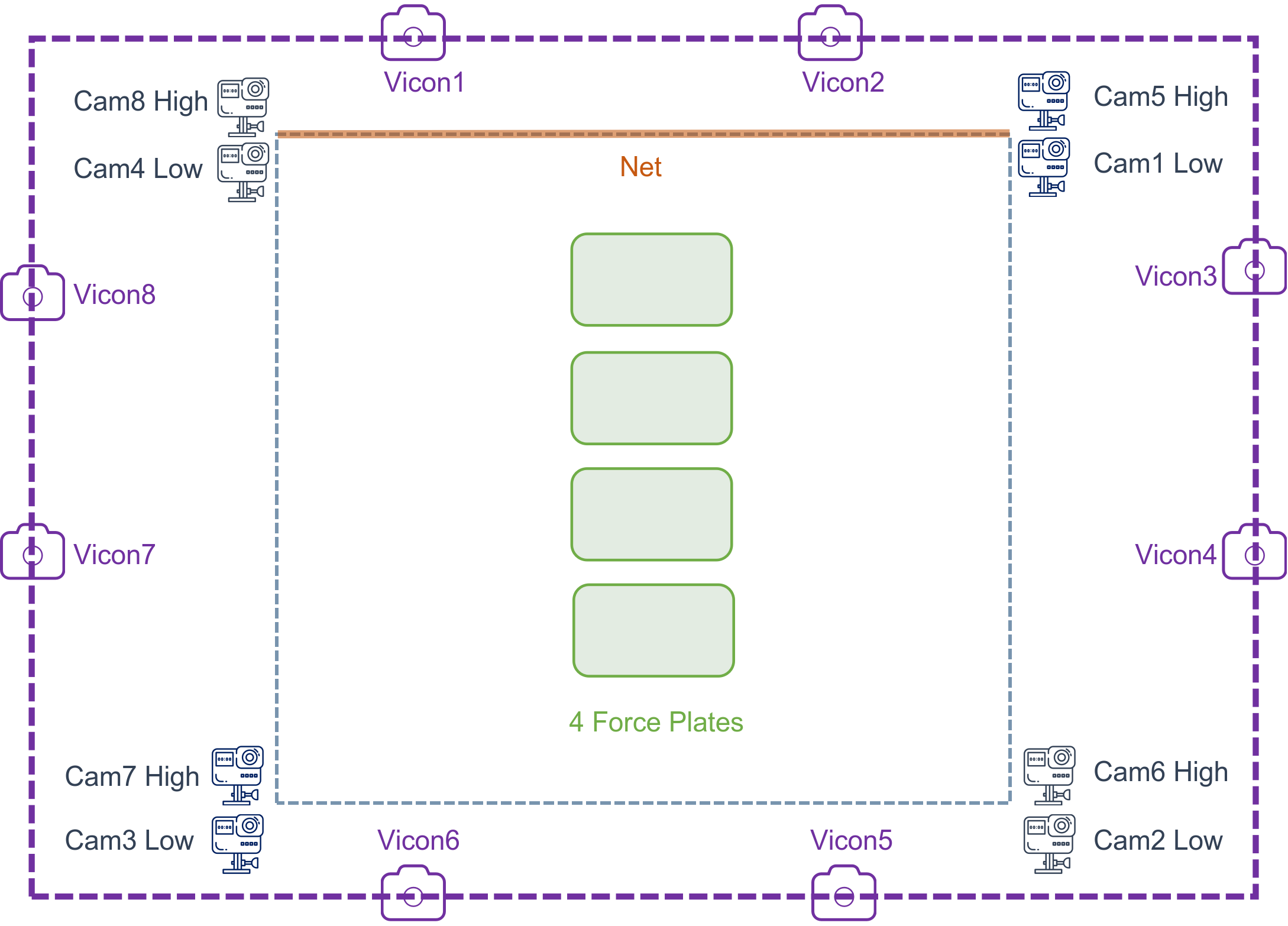}%
  \caption{Instrument geometry (plan view, schematic): four Kistler plates, eight static RGB viewpoints with independent streams, and Vicon coverage. Spatial layout only; temporal alignment is performed in post-processing (Sec.~\ref{sec:sync_grf}).}
  \label{fig:setup}
  \Description{Plan-view schematic of the instrumented court: force-plate placement, fixed multi-view camera layout, and mocap coverage (not timing).}
\end{figure}

\begin{figure*}[!t]
  \centering
  \includegraphics[width=\textwidth]{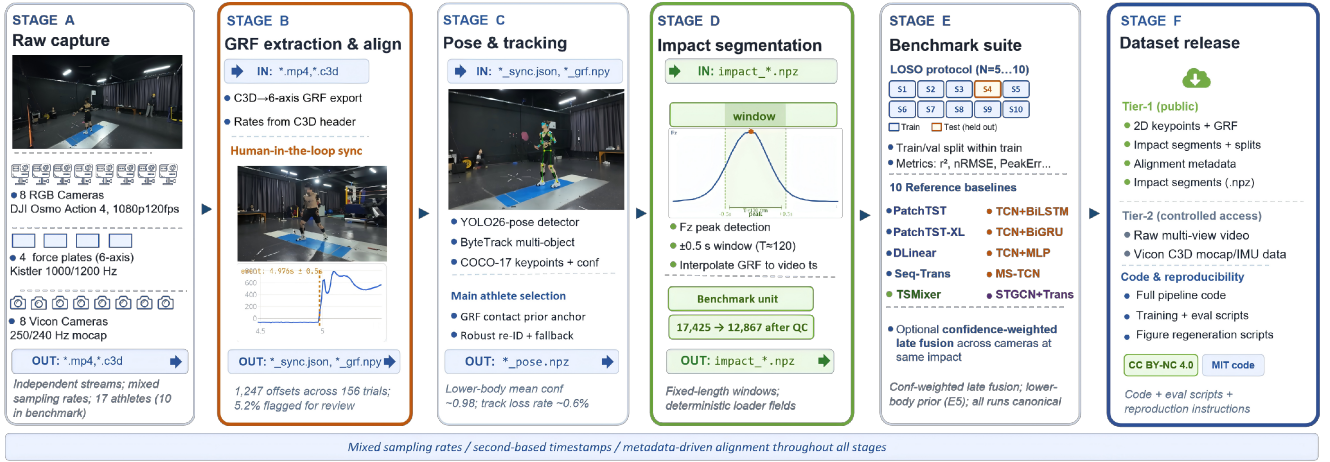}%
  \caption{Release pipeline (A--F, left to right): synchronized exports, GRF extraction and video--GRF alignment, pose estimation and tracking, fixed-duration impact segments, LOSO packaging with reference baselines and optional multi-view late fusion, and tiered distribution. All timing is represented on a continuous second axis to support mixed sample rates; court layout is illustrated in Fig.~\ref{fig:setup}.}
  \label{fig:pipeline}
  \Description{Six-stage horizontal pipeline (A--F): raw capture; GRF export and video--GRF alignment; pose and tracking; impact segments; LOSO benchmark with ten baselines and late fusion; tiered release.}
\end{figure*}

\noindent We trace the resource from capture to public benchmark. Fig.~\ref{fig:setup} sketches the instrument layout; Fig.~\ref{fig:pipeline} maps processing stages A--F. Subsections follow acquisition, synchronization, segmentation, statistics, and access policy.

\subsection{Acquisition Setup}
Recordings come from a fully instrumented court: up to eight fixed RGB cameras (DJI Osmo Action 4, \(\sim\)120 FPS), four Kistler six-axis force plates, and an eight-camera Vicon system exported to C3D. Vicon and plates share a laboratory time base; consumer RGB uses independent oscillators, so video--GRF alignment is established in software with human-annotated per-camera offsets (nominally within one frame) plus automated checks (Sec.~\ref{sec:sync_grf}) \cite{evansSynchronisedVideoMotion2024,werlingAddBiomechanicsDatasetCapturing2025}. Optional IMU streams are documented but omitted from the primary LOSO recipe because calibration drift varied across sessions.

\subsection{Subjects and Protocols}
Seventeen competitive badminton athletes (national second tier or higher; identifiers sub\_001--sub\_017) participated. The public benchmark tree uses ten subjects (eight men, two women), reflecting current recruitment constraints in elite programs as well as logistics for synchronized fatigue blocks; readers should treat gender and single-venue coverage as scope limits for subgroup claims. Archival RGB alone for those ten already exceeds \(1\,\mathrm{TB}\). All reported LOSO results use the packaged ten-subject splits.
Each athlete performs structured drills and match-like rallies before and after induced fatigue, with trial identifiers that encode stage and fatigue state for downstream analysis \cite{valldecabresEffectMatchFatigue2022,tongEffectsAnkleDorsiflexor2023,gaoAutomatedRecognitionAsymmetric2023,marchena-rodriguezIncidenceInjuriesAmateur2020}. The design cycles through rally play and three footwork modules (upper-net emphasis, smash-oriented patterns, six-point reactions), repeating the sequence after whole-court fatigue.
The protocol spans technically stable and workload-degraded conditions, which motivates careful RGB-to-GRF alignment as discussed in Sec.~\ref{sec:sync_grf}.

\subsection{Synchronization and Ground Truth GRF}
\label{sec:sync_grf}
\noindent\textbf{Relation to Fig.~\ref{fig:pipeline}.} Stage~B exports GRF from C3D, aligns video to GRF, and applies synchronization \QAabbrev{}; Stage~C runs pose estimation and tracking on time-corrected RGB.

\paragraph{Heterogeneous clocks and alignment scope.}
Vicon and plates share a single clock; each RGB camera free-runs and is not hardware-genlocked to the laboratory infrastructure---a frequent compromise in sports capture. We therefore estimate per-camera offsets in a dedicated annotation tool by pairing salient GRF and video events. Released metadata records offsets and optional uncertainty derived from annotation tolerance. Residual misalignment is bounded by the labeling procedure and by \(\pm0.5\,\mathrm{s}\) impact windows; interpolation policies appear in \BadmintonGRFSupplementary{}.

\paragraph{Ground-truth GRF and alignment target definition.}
Reference GRF is exported from the C3D pipeline by aggregating the four instrumented plates in the laboratory frame; vertical components follow a contact-positive \(F_z\) convention for supervision. The headline benchmark predicts \(F_z\) only, while \(F_x\) and \(F_y\) are retained for auxiliary studies because vertical loading remains the dominant injury surrogate \cite{lamBiomechanicsLowerLimb2020,pardiwalaBadmintonInjuriesElite2020,huCorrelationLowerLimb2022} and is less noisy under multi-plate stance. Mocap rates (\rateHz{240/250}) and plate rates (\rateHz{1000/1200}) are resampled on a shared continuous timeline in seconds rather than assuming a single frame grid.

\paragraph{Synchronization \QAabbrev{} and Stage~C processing.}
Automated \QAabbrev{} rules screen offset consistency; diagnostic plots are provided in \BadmintonGRFSupplementary{}. Among 156 packaged trials we recorded 1,247 $(\text{trial},\text{camera})$ offsets and escalated 65 (5.2\%) for manual reconciliation. Once accepted, Stage~C runs YOLO26-pose with ByteTrack \cite{chakrabarty2026YOLO26,zhangByteTrackMultiobjectTracking2022}; plate-derived contact cues help maintain identities when multiple players appear.

\noindent\emph{Interpretation.} Panel~A of Table~\ref{tab:qa_audit} degrades monotonically as synthetic timing errors grow from \(\pm1\) to \(\pm3\) frames, corroborating the intended one-frame alignment budget. Together with Panel~B, the audit suggests residual sync noise is smaller than cross-subject biomechanical variability at the current LOSO operating point, which is why Sec.~\ref{sec:experiments} reports four complementary metrics rather than a single score.

\subsection{Segment Construction and Data Statistics}
\label{sec:dataset_stats}
\paragraph{Impact segments.}
Stage~D ingests synchronized streams (Sec.~\ref{sec:sync_grf}), detects landing peaks on aligned \(F_z\), crops \(\pm0.5\,\mathrm{s}\) symmetric windows (\(T\approx120\) frames at nominal video rate), and writes one training instance per trial, camera, and detected peak with co-registered pose, metadata, and quality flags (Fig.~\ref{fig:pipeline}, Stage~D). Schema contracts live in \BadmintonGRFSupplementary{} and at \BadmintonGRFRepo{}. The reference loader keeps instances whose lower-body mean pose score is at least 0.70, whose tracking lost rate does not exceed 0.05, and whose peak \(F_z\) stays within three times body weight; inputs additionally apply low-score masking below 0.1, finite sanitization, and bounded clipping on positions and finite-difference derivatives.

\paragraph{Dataset statistics.}
The Tier~1 release contains 17{,}425 on-disk segments; loader gates yield 12{,}867 training instances covering 1{,}732 unique physical impacts after multi-view deduplication across 156 instrumented trials. Roughly 7\% of windows anchor on the manual sync event when automated peak search is ill-conditioned---a metadata flag that does not indicate sync failure. Stage histograms and optional LOSO bundles for $N\!\in\![5,10]$ subjects accompany \BadmintonGRFSupplementary{} and the \BadmintonGRFRepoShort{}. Table~\ref{tab:stats} condenses acquisition scope.

\begin{table}[t]
  \centering
  \footnotesize
  \setlength{\tabcolsep}{3pt}
  \renewcommand{\arraystretch}{1.03}
  \caption{Quality audits on Tier~1 segments. \textbf{A:} synthetic frame-wise misalignment stress test on the released pose-to-\(F_z\) pipeline (self-consistency; $n=17{,}425$). \textbf{B:} loader gates versus pre-gate inventory, including mean absolute frame lag between force peaks and pose references (details in \BadmintonGRFSupplementary{}).}
  \label{tab:qa_audit}
  \begin{tabular}{@{}rcccc@{}}
    \toprule
    \multicolumn{5}{@{}l@{}}{\textbf{A) Temporal-offset stress test}} \\
    \midrule
    $\Delta$ frame & $r^2\uparrow$ & RMSE (BW)$\downarrow$ & Peak err (BW)$\downarrow$ & Timing err (fr)$\downarrow$ \\
    \midrule
    $\pm 0$ & 1.000 & 0.000 & 0.000 & 0.00 \\
    $\pm 1$ & 0.951 & 0.133 & 0.534 & 1.03 \\
    $\pm 2$ & 0.869 & 0.220 & 0.923 & 2.03 \\
    $\pm 3$ & 0.779 & 0.286 & 1.094 & 3.02 \\
    \midrule
    \multicolumn{5}{@{}l@{}}{\textbf{B) Loader-gate audit}} \\
    \midrule
    \multicolumn{1}{@{}l}{Split} & \multicolumn{1}{c}{Segments} & \multicolumn{1}{c}{Unique impacts} & \multicolumn{2}{c@{}}{Mean $|\mathrm{lag}|$ (fr)} \\
    \midrule
    \multicolumn{1}{@{}l}{Before gates} & \multicolumn{1}{c}{17{,}425} & \multicolumn{1}{c}{2{,}263} & \multicolumn{2}{c@{}}{15.89} \\
    \multicolumn{1}{@{}l}{After gates} & \multicolumn{1}{c}{12{,}867} & \multicolumn{1}{c}{1{,}732} & \multicolumn{2}{c@{}}{15.71} \\
    \bottomrule
  \end{tabular}
\end{table}
\noindent Table~\ref{tab:qa_audit} quantifies sensitivity to timing jitter (Panel~A perturbs alignment by integer frame shifts in a self-consistency study) and to loader gates. Panel~B lists mean absolute frame lag between each \(|F_z|\) peak and its pose-derived reference instant, averaged over the release; formal definitions accompany \BadmintonGRFSupplementary{}. After gating, 73.8\% of segments remain; the largest protocol share shift is 13.05 percentage points versus 2.21 points by subject, so stage imbalances deserve explicit reporting in stratified analyses.

\begin{table}[t]
  \centering
  \small
  \setlength{\tabcolsep}{4pt}
  \renewcommand{\arraystretch}{1.05}
  \caption{Benchmark scope and sensing summary for the public Tier~1 LOSO subset (ten subjects, processed segments and metadata).}
  \label{tab:stats}
  \begin{tabularx}{\columnwidth}{@{}lX@{}}
    \toprule
    \textbf{Item} & \textbf{Value} \\
    \midrule
    Subjects (benchmark) & 10 (sub\_001--sub\_010) \\
    Subjects (collected) & 17 (complete; ongoing collection) \\
    Cameras & 8 DJI Osmo Action 4, 1080p, \(\sim\)120 FPS \\
    Force plates & 4 Kistler 9260AA6 plates, 6-axis GRF \\
    Force sampling rate & \rateHz{1000/1200} (\(=4\!\times\)\rateHz{250/240} mocap; per-trial metadata) \\
    Mocap & 8 Vicon T40 cameras, \(\sim\)52-marker protocol (14 mm; session counts vary slightly), C3D (raw markers preserved) \\
    Mocap sampling rate & \rateHz{240/250} (\(=1/4\) force rate; per-trial metadata) \\
    Impact segments & 17{,}425 archived segments; 12{,}867 after quality gates; 1{,}732 unique impacts (post-gate multi-view dedup.); 156 trials \\
    Segment window & \(\pm 0.5\) s, \(T\approx 120\) frames \\
    \bottomrule
  \end{tabularx}
\end{table}

\subsection{Data Access and Privacy}
\label{sec:data_access}
Tier~1 publicly distributes processed pose, aligned GRF, metadata, and predefined splits.
The Zenodo record is \url{https://doi.org/10.5281/zenodo.19277566}.
During peer review we enable access to the restricted Tier~1 deposit on request (anonymous requests supported), typically within two to three business days.
Tier~2 grants raw RGB and C3D to approved applicants under agreements that forbid redistribution and re-identification.
The CC BY-NC~4.0 license on Tier~1 reflects athlete consent and host-institution policies on redistributable derived signals (pose and processed GRF); it is not intended to restrict non-commercial academic use of the benchmark.
Further documentation resides in \BadmintonGRFSupplementary{} and on the \BadmintonGRFRepoShort{}.
Replicating this instrumentation and athlete cohort outside supported laboratories is demanding; the release is therefore intended as a community benchmark for multimodal sports biomechanics.

\section{Benchmark and Evaluation}
\label{sec:benchmark_methods}\label{sec:experiments}
\noindent We stress \emph{resource utility} (Fig.~\ref{fig:pipeline}, Stage~E): reference models calibrate difficulty rather than claim state of the art. Table~\ref{tab:benchmark_all} reports \textsc{LOSO} single-view (SV) scores first, followed by \textsc{LOSO} late fusion (Fus) and Within-trial (Within) diagnostics; SV is the primary cross-subject comparison. Every model shares frozen LOSO manifests, a 15\% validation carve-out inside training subjects, and best-on-validation checkpoints so comparisons do not depend on ad hoc splits. Hyperparameters, fold-wise dispersion, confidence intervals, and paired tests appear in \BadmintonGRFSupplementary{}.

\begin{table*}[!t]
  \centering
  \footnotesize
  \setlength{\tabcolsep}{2.4pt}
  \renewcommand{\arraystretch}{1.12}
  \caption{\textbf{Reference benchmark} (macro means over folds): \textsc{LOSO} SV/Fus and Within SV/Fus. Bold = best per block; favor higher \(r^2\) and lower \textbf{R}/\textbf{P}/\textbf{T} (definitions in the table legend). Leading \textsc{LOSO} SV scores overlap within cross-fold variability---read PatchTST and ST-GCN+Transformer jointly. Per-fold tables ship with the release (\BadmintonGRFRepoShort{}).}
  \label{tab:benchmark_all}
  \resizebox{\textwidth}{!}{%
  \begin{tabular}{@{}l cccc cccc cccc cccc@{}}
    \toprule
    & \multicolumn{4}{c}{\textbf{\textsc{LOSO} SV}} & \multicolumn{4}{c}{\textbf{\textsc{LOSO} Fus}} & \multicolumn{4}{c}{\textbf{Within SV}} & \multicolumn{4}{c}{\textbf{Within Fus}} \\
    \cmidrule(lr){2-5} \cmidrule(lr){6-9} \cmidrule(lr){10-13} \cmidrule(lr){14-17}
    \textbf{Model} & \(r^2\,\uparrow\) & \textbf{R}\,\(\downarrow\) & \textbf{P}\,\(\downarrow\) & \textbf{T}\,\(\downarrow\) & \(r^2\,\uparrow\) & \textbf{R}\,\(\downarrow\) & \textbf{P}\,\(\downarrow\) & \textbf{T}\,\(\downarrow\) & \(r^2\,\uparrow\) & \textbf{R}\,\(\downarrow\) & \textbf{P}\,\(\downarrow\) & \textbf{T}\,\(\downarrow\) & \(r^2\,\uparrow\) & \textbf{R}\,\(\downarrow\) & \textbf{P}\,\(\downarrow\) & \textbf{T}\,\(\downarrow\) \\
    \multicolumn{17}{@{}c}{\scriptsize \textbf{Metrics:} \(r^2\) = coefficient of determination; \textbf{R} = RMSE \(F_z\) (BW); \textbf{P} = peak magnitude error (BW); \textbf{T} = peak timing error (frames).} \\
    \midrule
    PatchTST~\cite{nieTimeSeriesWorth2023}                     & \textbf{0.403} & \textbf{0.510} & 0.226 & 1.07 & 0.412 & 0.584 & 0.693 & 1.48 & 0.494 & 0.470 & 0.211 & 1.33 & 0.304 & 0.634 & 0.719 & 3.98 \\
    ST-GCN+Transformer~\cite{shiSkeletonBasedActionRecognition2019,shiTwoStreamAdaptiveGraph2019,vaswaniAttention2017}           & 0.394 & 0.514 & 0.221 & \textbf{0.96} & 0.454 & \textbf{0.487} & \textbf{0.210} & \textbf{0.90} & 0.576 & 0.430 & \textbf{0.191} & \textbf{1.27} & \textbf{0.617} & \textbf{0.408} & \textbf{0.182} & \textbf{1.15} \\
    TCN+BiGRU~\cite{baiEmpiricalEvaluationConvolutional2018,choLearningPhraseRepresentations2014}                    & 0.390 & 0.514 & 0.348 & 3.79 & 0.274 & 0.645 & 0.952 & 6.51 & 0.519 & 0.456 & 0.407 & 4.27 & 0.167 & 0.690 & 1.025 & 8.99 \\
    TCN+BiLSTM~\cite{baiEmpiricalEvaluationConvolutional2018,hochreiterLongShortTermMemory1997}                   & 0.374 & 0.521 & 0.309 & 2.58 & 0.370 & 0.602 & 0.960 & 3.73 & 0.547 & 0.443 & 0.365 & 3.96 & 0.151 & 0.696 & 1.021 & 8.93 \\
    TSMixer~\cite{chenTSMixerAllMLP2023}                      & 0.351 & 0.531 & \textbf{0.218} & 1.85 & \textbf{0.472} & 0.553 & 0.673 & 1.22 & \textbf{0.635} & \textbf{0.395} & 0.201 & 1.67 & 0.414 & 0.579 & 0.645 & 2.11 \\
    Seq-Transformer~\cite{vaswaniAttention2017}              & 0.345 & 0.533 & 0.281 & 1.82 & 0.336 & 0.619 & 0.870 & 2.37 & 0.517 & 0.458 & 0.245 & 1.69 & 0.032 & 0.745 & 0.900 & 5.36 \\
    PatchTST-XL~\cite{nieTimeSeriesWorth2023}                  & 0.339 & 0.535 & 0.224 & 1.46 & 0.456 & 0.561 & 0.656 & 1.38 & 0.597 & 0.417 & 0.216 & 1.45 & 0.374 & 0.600 & 0.713 & 2.58 \\
    TCN+MLP~\cite{baiEmpiricalEvaluationConvolutional2018}                      & 0.275 & 0.562 & 0.432 & 9.29 & 0.327 & 0.625 & 1.103 & 9.47 & 0.341 & 0.537 & 0.587 & 12.68 & 0.107 & 0.716 & 1.055 & 13.89 \\
    MS-TCN~\cite{farhaMSTCN2019}                       & 0.184 & 0.596 & 0.688 & 14.24 & 0.232 & 0.667 & 1.313 & 13.92 & 0.122 & 0.621 & 0.764 & 15.34 & 0.135 & 0.708 & 1.357 & 15.11 \\
    DLinear~\cite{zengAreTransformersEffective2023}                      & 0.072 & 0.636 & 0.819 & 16.03 & 0.149 & 0.702 & 1.432 & 15.11 & 0.003 & 0.662 & 0.939 & 15.96 & 0.021 & 0.753 & 1.532 & 15.09 \\
    \bottomrule
  \end{tabular}%
  }
  \vspace{2pt}
  {\scriptsize\setlength{\baselineskip}{10.4pt}%
  \noindent\emph{Reproducibility:} frozen training bundles and bundle identifiers are listed in \BadmintonGRFSupplementary{}.}
\end{table*}

\paragraph{Task, models, and metrics.}
The Tier~1 \textbf{keypoint-first} task regresses BW-normalized \(F_z\) from COCO-17 joints, confidences, and finite-difference motion features \cite{mundt2022Estimating,ishida2024Estimation,hossain2025Knowledge} given known impact alignment (Sec.~\ref{sec:dataset_stats}). We evaluate ten architectures spanning temporal convolutions, recurrent models, transformers, and skeleton graph networks \cite{baiEmpiricalEvaluationConvolutional2018,hochreiterLongShortTermMemory1997,choLearningPhraseRepresentations2014,vaswaniAttention2017,zengAreTransformersEffective2023,nieTimeSeriesWorth2023,chenTSMixerAllMLP2023,farhaMSTCN2019,shiSkeletonBasedActionRecognition2019,shiTwoStreamAdaptiveGraph2019}. Global \(r^2\) and RMSE summarize curve fit, while peak and timing errors focus on landing kinetics; confidence-weighted late fusion can raise \(r^2\) yet hurt peak metrics, so we discourage single-score rankings.

\paragraph{Evaluation settings and how to read Table~\ref{tab:benchmark_all}.}
\textbf{(i)~SV} holds out entire subjects; \textbf{(ii)~Fus} averages camera-specific predictions with confidence weights under the same LOSO folds; \textbf{(iii)~Within} holds out impacts inside seen trials as a high-bias diagnostic. Training always reserves 15\% of held-in subjects for validation before selecting checkpoints. PatchTST and ST-GCN+Transformer lead SV with \(r^2=0.403/0.394\), yet paired fold-level comparisons on the archived LOSO bundle do not show a stable separation; spread across subjects dominates residual architecture gaps. Macro means aggregate all protocols; fatigue- or stage-stratified results, LOSO curves for $N\!\in\![5,10]$, and camera ablations are deferred to \BadmintonGRFSupplementary{}.

\paragraph{Reading Table~\ref{tab:benchmark_all}.}
\textsc{LOSO} SV clusters near \(r^2\approx0.36\)--\(0.40\), aligned with Table~\ref{tab:qa_audit}: subject shift dominates over sync noise. PatchTST leads \(r^2\)/RMSE; ST-GCN+Transformer leads peak magnitude and timing (\textbf{P}/\textbf{T}), plausibly from explicit graph structure on short landing transients. PatchTST-XL trails PatchTST; TCN+MLP, MS-TCN, and DLinear show very large \textbf{T}, so timing error differentiates weak baselines. Late fusion can increase \(r^2\) (TSMixer \(0.472\)) but often hurts \textbf{R}/\textbf{P} when per-view peaks conflict (PatchTST \(P\) rises from \(0.226\) to \(0.693\)); ST-GCN+Fus is strongest on fused \textbf{R}/\textbf{P}/\textbf{T}, whereas recurrent TCN stacks degrade under fusion. Within-trial \(r^2\) reaches \(\sim\!0.6\), isolating subject shift; fusion helps some models (ST-GCN) but not others (PatchTST). We therefore stress all four metrics jointly rather than a single rank.

\paragraph{Applications and artifact release.}
Downstream uses include sports workload analytics, fatigue-conditioned modeling, and multi-view learning under occlusion \cite{johnsonOnfieldPlayerWorkload2019,gaoAutomatedRecognitionAsymmetric2023,mundt2022Estimating,bogaertPredictingVerticalGround2024}. Tier~1 modalities are available under CC BY-NC~4.0, while companion software uses MIT licensing; Tier~2 media follows the access policy linked from the \BadmintonGRFRepoShort{}. Reporting choices follow community guidance for biomechanical datasets \cite{hebert-losierReportingGuidelinesRunning2023}.

\section{Discussion, Reproducibility, and Limitations}
\label{sec:limitations}

\paragraph{Reproducibility protocol.}
Alongside Tier~1 data we publish LOSO manifests, deterministic segment lists after gating, random seeds, optimizer presets, command-line recipes, canonical metric scripts, and the synchronization and filter audits underlying Table~\ref{tab:qa_audit}. Together they separate modeling choices from dataset ambiguity and ease cross-site replication; line-level manifests remain in \BadmintonGRFSupplementary{}.

\paragraph{Metric interpretation and reporting.}
Landing impulses concentrate energy in short temporal supports, so curve smoothers can lower RMSE while blunting peaks, whereas sharper regressors may trade timing accuracy. We therefore ask users to quote \(r^2\), RMSE, peak magnitude error, and peak timing error jointly, alongside fold means and standard deviations, post-gate cohort sizes, fusion-weight policies, and masking schemes for low-confidence joints.

\paragraph{Metadata-enabled diagnostics.}
Trial identifiers embed stage and fatigue metadata so analysts can stratify without altering loaders---useful for probing protocol shifts between fresh and fatigued captures.

\paragraph{Practical reuse guidance.}
Document four decisions whenever numbers are reported: gated versus full exports, masking prior to temporal derivatives, globally fixed versus fold-tuned fusion weights, and macro averages versus stage/fatigue slices. Internal pilots showed each lever can move RMSE or peak scores without changing the backbone, underscoring that preprocessing transparency matters as much as model class for impact regression.

\paragraph{Summary and limitations.}
The present LOSO ceiling near \(r^2\approx0.40\) tracks cross-subject biomechanics more than residual synchronization: Table~\ref{tab:qa_audit} Panel~A implies roughly \(\Delta r^2\approx0.05\) per adversarial frame shift, whereas held-out subjects induce \(>\!0.15\) swings. Uncalibrated RGB still demands software offsets with documented uncertainty (Sec.~\ref{sec:sync_grf}), yet windows of \(\pm0.5\,\mathrm{s}\) bound residual phase error. Future gains likely require subject-aware or metadata-conditioned objectives rather than marginal alignment tweaks.

\paragraph{Coverage, access, and ethics.}
The open tree currently spans ten athletes at a single site; completing \QAabbrev{} on the remaining instrumented athletes will unlock broader LOSO sweeps (\BadmintonGRFSupplementary{} summarizes \(N\!\in\![5,10]\) pilots). Tier~1 suffices to reproduce Table~\ref{tab:benchmark_all}; Tier~2 unlocks appearance and full mocap studies. Institutional review board approval and tiered informed consent govern capture; extended ethics discussion appears in \BadmintonGRFSupplementary{}.

\paragraph{Supplementary structure and frozen exports.}
\BadmintonGRFSupplementary{} (Secs.~A--H) hosts schema tables, training defaults, statistical appendices, and optional machine-readable audit bundles referenced throughout the paper; it tracks the public GitHub release. Table~\ref{tab:benchmark_all} was verified against the frozen export archived with that release.

\paragraph{Failure modes and extensions.}
Residual errors cluster around heavy occlusions, identity switches during crossings, and borderline impacts near gating thresholds. Metadata supports targeted fixes such as uncertainty-aware objectives, fatigue-conditioned heads, or geometry-informed fusion. Expanding QA to the remaining subjects will tighten confidence intervals for subgroup analyses.

\paragraph{Broad impact beyond biomechanics.}
For multimedia research the corpus offers stress tests on occlusion, viewpoint diversity, and pose-to-dynamics transfer from laboratory calibration to training-hall conditions. For embodied systems, synchronized pose and force traces can supervise contact-aware planning and imitation in agile sports agents. Throughout, we treat synchronization QA, deterministic filtering, and multi-metric reporting as integral to the resource rather than optional footnotes.

\paragraph{Closing remarks.}
BadmintonGRF connects intermittent-court motion, audited heterogeneous multimodal timing, and a specified benchmark so researchers in multimedia and biomechanics can share a common supervision signal instead of bespoke splits.
We anticipate Tier~1 will lower the entry cost for privacy-conscious pose-to-force studies, while Tier~2 enables richer appearance- and marker-based modeling when controlled access is appropriate; both tiers are documented with the same reproducibility metadata.

\begin{acks}
We thank the athletes and staff who supported instrumented data collection, and colleagues who provided feedback on releases and documentation.
This work was supported in part by the National High-Level Talent Special Support Program (Ten Thousand Talents Program) for Young Talents (Grant No.~588020-X42506); the Key Research and Development Program of Zhejiang Province (Grant No.~2024SSYS0026); the Fundamental Research Funds for the Central Universities; and the Beijing Natural Science Foundation (Grant No.~4262068).
\end{acks}

\balance
\bibliographystyle{ACM-Reference-Format}
\bibliography{ACM_MM_2026}

\end{document}